\documentclass[letterpaper, 10 pt, conference]{ieeeconf}  

\IEEEoverridecommandlockouts                              

\usepackage{graphics} 
\usepackage{epsfig} 
\usepackage{amsmath} 
\usepackage{amssymb}  
\usepackage{algorithm,algorithmic}

\usepackage[T1]{fontenc}

\title{\LARGE \bf
Quad2Plane: An Intermediate Training Procedure for Online Exploration in Aerial Robotics via Receding Horizon Control
}

\author{Alexander Quessy$^{1}$ and Thomas Richardson$^{2}$
\thanks{$^{1}$Alexander Quessy is with Department of Aerospace Engineering, Queen's Building, University of Bristol, University Walk, Bristol BS8 1TR
        {\tt\small aq15777@bristol.ac.uk}, \textit{corresponding author}}%
\thanks{$^{2}$Thomas Richardson is with Department of Aerospace Engineering, Queen's Building, University of Bristol, University Walk, Bristol BS8 1TR
        {\tt\small thomas.richardson@bristol.ac.uk}}%
}

\begin{document}

\maketitle
\thispagestyle{empty}
\pagestyle{empty}

\begin{abstract}

Data driven robotics relies upon accurate real-world representations to learn useful policies. Despite our best-efforts, zero-shot sim-to-real transfer is still an unsolved problem, and we often need to allow our agents to explore online to learn useful policies for a given task. For many applications of field robotics online exploration is prohibitively expensive and dangerous, this is especially true in fixed-wing aerial robotics. To address these challenges we offer an intermediary solution for learning in field robotics. We investigate the use of dissimilar platform vehicle for learning and offer a procedure to mimic the behavior of one vehicle with another. We specifically consider the problem of training fixed-wing aircraft, an expensive and dangerous vehicle type, using a multi-rotor host platform. Using a Model Predictive Control approach, we design a controller capable of mimicking another vehicles behavior in both simulation and the real-world.

\end{abstract}

\section{INTRODUCTION}

In the past decade Deep Reinforcement Learning has proven to be an effective tool to solve a variety of simulated \cite{IROS1}\cite{IROS2} and real-world \cite{IROS3}\cite{IROS4} control tasks. Real-world robotic learning is plagued by a host of practical issues \cite{IROS41} such as: resets, state-estimation and platform integrity. These practical limitations are easily solved by learning in simulation, but this often leads to a \textit{reality gap}, where the policy learned by the agent learns to exploit attributes that are not present in the real-world, leading to poor performance. A common robotic learning pipeline is therefore to complete the majority of training in simulation, where demonstrations are cheap, and then fine-tune the learned policy on a real-world robot. 

This approach works well in robotic domains where the vehicle can safely explore the environment without significant constraints, such as gripper \cite{IROS5} robots. However, in many field-robotic settings it is not safe to allow the robot to explore and fine-tune a policy in the real-world, as the vehicle can damage itself and 3rd parties. This is a challenge that is particularly constraining in fixed-wing aerial robotics, where the platform is often large and financially expensive. Further, we are often most interested in fine-tuning policies when the aircraft is slow and close to the ground \cite{IROS6} \cite{IROS47}, where it is most likely to cause damage. This makes real-world fixed-wing robotic learning difficult constraining most research to simulation \cite{IROS7}. 

In comparison, multi-rotor drones are ubiquitous in aerial robotic research \cite{IROS34}, and have been used for online robotic learning in a variety of safe \cite{IROS36} and constrained real-world environments \cite{IROS35}. Fixed-wing aircraft typically have superior range, endurance and load-carrying capabilities than rotary aircraft. However, this is not a constraint on the majority of learning tasks, because most stochastic control problems encountered by the robot, where using a DRL based controller would be useful, occur at the beginning or end of a flight and carrying a payload does not effect the learning process. Along with being safer, multi-rotors offer several distinct advantages, compared to fixed-wing aircraft, for online learning:
\begin{itemize}
        \item Multi-rotors can hover and translate linearly in all three flight-axis, making it easier to reset the learning process online. \cite{IROS8}.
        \item Multi-rotors are not as geometrically constrained as fixed-wing aircraft, allowing for easier installation of compute devices and sensors. This makes it easier to get accurate state-estimation and to perform online policy optimization \cite{IROS9}. 
        \item Multi-rotors have a larger low-speed flight envelope, easily being able to hover and navigate tight, constrained environments\cite{IROS10}. This allows us to train the robot in scenarios closer to the platforms limits without the risk of crashing, opening the door for safe-RL \cite{IROS19} and lifelong learning \cite{IROS18}. 
\end{itemize}

Given how much easier it is to learn control policies on multi-rotors than fixed-wing aircraft: \textit{can a multi-rotor mimic the mechanics of a fixed-wing aircraft}? This would allow us to learn useful policies in the real-world, without the burden of reset-ability and safety imposed by a fixed-wing aircraft. All whilst providing access to the underlying real-world state-observation distribution, including wind and imaging data, which is difficult to entirely capture in simulation \cite{IROS37}.

In this paper we present a general procedure to mimic the dynamics of a \textit{target} vehicle on a dissimilar \textit{platform} vehicle. Specifically we consider the problem of replicating the dynamics of a fixed-wing target vehicle on a multi-rotor platform. We make the following contributions:
\begin{itemize}
        \item We pose the general platform-to-target control problem as a discrete time dynamic problem in section \ref{sec:problem_setting}. We then design an optimal model based controller to solve this problem in section \ref{sec:mpc}.
        \item In section \ref{sec:implementation} we provide a description of the implementation of our controller for aerial robotics, along with the rationale behind the design of our vehicles state based cost function. This offers useful insights into how we need to design vehicles that are used for robotic learning.
        \item In section \ref{sec:experiments} we use a combination of simulated and real-world data to validate our approach and describe the key limitations of our approach.
\end{itemize}

\section{RELATED WORK}

The problem of transferring learned skills from simulation to the real-world is well documented within the robotic learning community \cite{IROS20}\cite{IROS21}\cite{IROS22}. This \textit{reality gap} is caused by a distributional shift from simulated to real-world observations, due to un-modelled physical effects such as noise in a collected image or friction in an actuator. One approach to solve this problem is to vary the state distribution of the simulation, either by randomizing the domain \cite{IROS11}\cite{IROS15} or dynamics \cite{IROS14}. However, Domain Randomization (DR) has some key limitations: the creation of suitable simulated training environments is time consuming, and deciding what to randomize requires care to ensure the final distributions match. Additionally there is no guarantee that DR can achieve zero-shot sim-to-real transfer \cite{IROS40}.

Alternatively, demonstrations can be used to help fine-tune the policy \cite{IROS17}\cite{IROS12}\cite{IROS23}, but for many settings it is difficult to obtain expert demonstrations to learn from. Learning in the real-world is perhaps the simplest way to ensure we don't have a reality gap, either by fine-tuning \cite{IROS24}, or without simulation altogether \cite{IROS16}. Unfortunately this requires our robotic platform to be capable of safe exploration, something that cannot be assured for our target vehicle. We offer an intermediary solution to this, by learning on a surrogate vehicle, we contribute a procedure to train inherently dangerous vehicles in the real-world safely.

Mimicking the dynamics of one vehicle with another has been used by the aerospace community for flight crew training \cite{IROS32} and aircraft flying quality evaluation \cite{IROS31} since the late 1950s \cite{IROS25}. These In Flight Trainer (IFT) aircraft \cite{IROS26} \cite{IROS33} typically use linear state-space model reference adaptive controllers based on $\mathcal{L}_{1}$ or LQR to mimic another aircraft's flight mechanics. This makes sense for IFT aircraft, where the objective is typically to provide proprioceptive information to the flight crew. However, it is desirable for our system to also be capable of replicating the vehicles non-linear dynamics, linear state-space control is therefore inappropriate. 

Model Predictive Control (MPC) is an effective method to control non-linear systems \cite{IROS38}, especially when constraints are imposed on exploration \cite{IROS49}. Classical MPC provides a complete trajectory to solve an optimal control problem, for complex optimization functions, as is experienced in non-linear control, this is often computationally intractable. Receding-horizon differential dynamic programming \cite{IROS27}, helps to address this by optimizing over a shorter (receding) horizon, rather than the full control time of the process. In this research we design a non-linear receding model predictive controller \cite{IROS48}, and find this is a procedure capable of mimicking the dynamics of our target fixed-wing vehicle, provided sufficient compute, time-horizon and model accuracy.



\section{PROBLEM SETTING}
\label{sec:problem_setting}
Whilst the primary focus of our work is in matching the dynamics of dissimilar aircraft types, the procedure we develop is applicable to many other robotic learning tasks. Consider a discrete time dynamic system as in

\begin{equation}
        \mathbf{x}_{t+1} = \mathbf{F}^{i}(\mathbf{x}_{t}, \mathbf{u}_{t})
        \label{eq:A-Discrete-Time}
\end{equation}

\begin{figure}[htb!]
        \centering
        \includegraphics[width=0.4\textwidth]{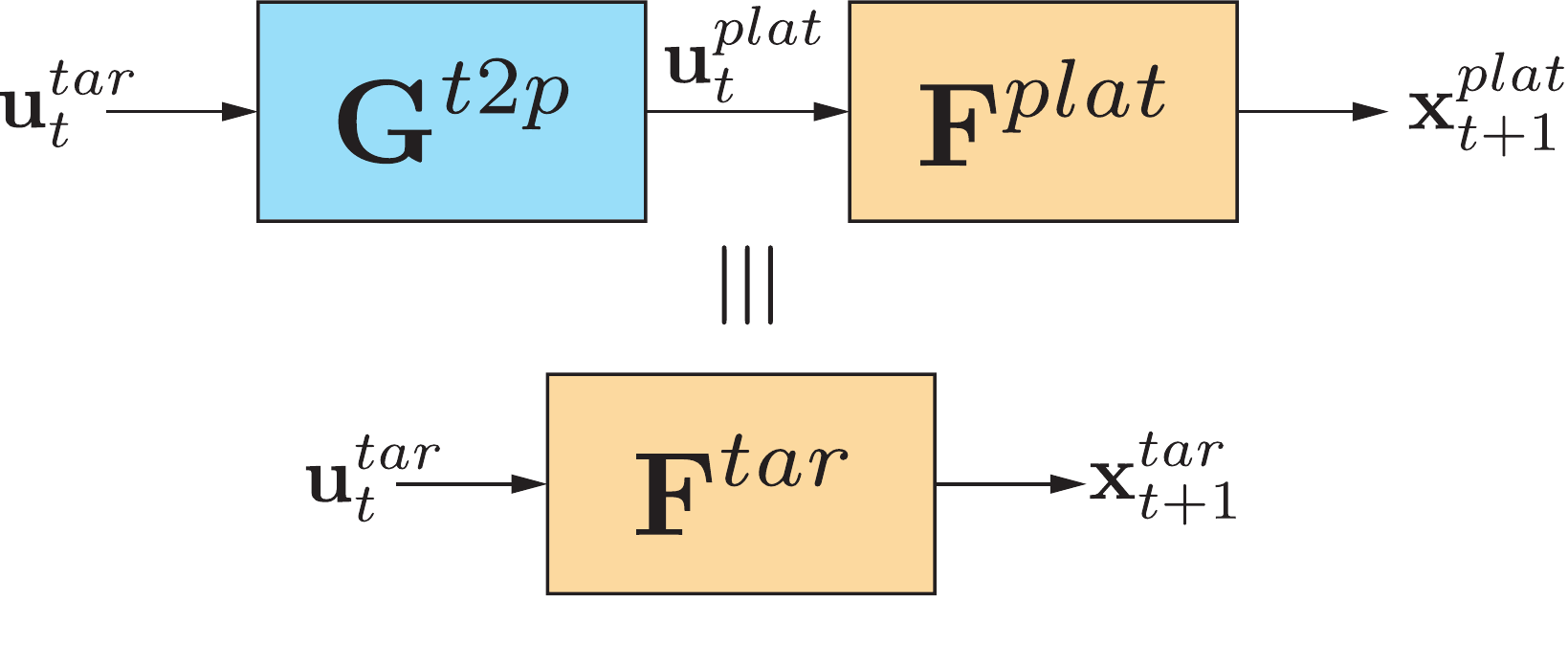}
        \caption{Problem Formulation Block Diagram}
        \label{fig:A-Problem_Block}
\end{figure}

Where $\mathbf{x}_{t} \in \mathbb{R}^{n}$ is the state of a vehicle $i$ at time $t$, $\mathbf{u} \in \mathbb{R}^{m}$ is the input to the vehicle at time $t$, and $\mathbf{F}$ is the vehicles state-transition. Under this framework we consider target and platform vehicles with transition functions $\mathbf{F}^{tar}$ and $\mathbf{F}^{plat}$ respectively. Our objective is therefore to design a controller $\mathbf{G}^{t2p}$, as in (\ref{eq:B-controller}), so that when $\mathbf{F}^{plat}$ receives a control input $\mathbf{u}_{t}^{plat}$ it undergoes the same state-transition as $\mathbf{F}^{tar}$, receiving an input $\mathbf{u}_{t}^{tar}$, as shown in figure \ref{fig:A-Problem_Block}.

\begin{equation}
        \mathbf{u}_{t}^{plat} = \mathbf{G}^{t2p}(\mathbf{u}_{t}^{tar}; \mathbf{x}_{t}^{plat})
        \label{eq:B-controller}
\end{equation}

We assess the performance of $G^{t2p}$ using a loss function between the target and platform trajectory terms $S_{\tau} = \mathcal{L}(\tau_{\mathbf{x}^{plat}}, \tau_{\mathbf{x}^{tar}})$.


\section{MODEL PREDICTIVE CONTROL}
\label{sec:mpc}
We can frame the control design problem for $G^{t2p}$ as an optimal control problem, for which receding horizon MPC offers a useful framework. We assume that the transition of the vehicles dynamics $\mathbf{F}$ is deterministic, and the model representation entirely captures the vehicles dynamics. 

Consider the cost function $\mathcal{J}(\mathbf{u}; \mathbf{x})$, composed of state $\mathbf{q}(\mathbf{x})$ and control $\mathbf{p}(\mathbf{u})$ dependant costs, as in

\begin{equation}
        \mathcal{J}(\mathbf{u}; \mathbf{x}) = \mathbf{q}(\mathbf{x}) + \mathbf{p}(\mathbf{u}) \ s.t. \ \mathbf{u}_{min} \le \mathbf{u} \le \mathbf{u}_{max}
        \label{eq:C-Cost_Function}
\end{equation}

The objective is then to find the optimal control sequence $\mathbf{u}^{*}$ that minimizes $\mathcal{J}$ over a finite time-horizon $T$, subject to the linear bounding constraint $[\mathbf{u}_{min}, \mathbf{u}_{max}]$. To help reduce the computational cost at each time $t$ the optimization problem is \textit{warm-started} with the previous control sequences solution, reducing the computational cost of the online optimization problem. 

Control sequence $\tau_{\mathbf{u}}^{tar}$ with finite length $T$ is the control input sequence received by $G^{t2p}$. This is the command sequence produced by the outer controller we are aiming to train. In a typical model-free RL setting only a single next time-step would be provided and the MPC time-horizon would effectively be one step. We found that this often causes instability in our platform vehicle as the MPC controller is unable to solve the non-linear control problem, effectively getting stuck in a local optima. To learn an RL policy it is therefore necessary to rollout a control trajectory $T$ steps ahead. When the policy is deployed online only the first state from the policies action trajectory will be selected.

Algorithm \ref{algo:A-t2p_MPC}, is a general MPC based procedure to mimic the dynamics of a target vehicle.. We denote the target vehicle model and platform model as $\hat{\mathbf{F}}^{tar}$ and $\hat{\mathbf{F}}$ respectively.

\begin{algorithm}[H]
        \caption{t2p-MPC}
        \begin{algorithmic}[1]
                \renewcommand{\algorithmicrequire}{\textbf{Input:}}
                \renewcommand{\algorithmicensure}{\textbf{Output:}}
                \REQUIRE $\tau_{\mathbf{u}^{tar}}$: target control input sequence, $\tau_{\mathbf{u}}$: last control prediction sequence (up to $T-1$)
                \ENSURE $\mathbf{u}_{t+1}$: next control input to platform
                \STATE $\tau_{\mathbf{x}^{tar}} \gets \emptyset$
                \FOR {$t$ to $T$}
                        \STATE $\hat{\mathbf{x}}_{t+1}^{tar} \gets \hat{\mathbf{F}}^{tar}(\mathbf{x}_{t}; \mathbf{u}_{t}^{tar})$; add to $\tau_{\mathbf{x}^{tar}}$
                \ENDFOR
                \STATE $\mathcal{J} \gets 0$
                \WHILE {not converged}
                        \FOR {$t$ to $T$}
                                \STATE $\hat{\mathbf{x}}_{t} \gets \hat{\mathbf{F}}(\mathbf{x}_{t}; \mathbf{u}_{t})$
                                \STATE $j \gets \mathbf{q}(\mathbf{\hat{x}}_{t}) + \mathbf{p}(\mathbf{u}_{t})$
                                \STATE $\mathcal{J} \gets \mathcal{J} + j$ 
                        \ENDFOR
                \STATE $\tau_{\mathbf{u}}$ $\gets$ $\underset{\mathbf{u}}{\mathrm{argmin}} \ \mathcal{J} \ s.t. \ \mathbf{u}_{min} \le \mathbf{u} \le \mathbf{u}_{max}$
                \ENDWHILE
        \end{algorithmic}
        \label{algo:A-t2p_MPC}
\end{algorithm}


\section{IMPLEMENTATION}
\label{sec:implementation}
\subsection{Simulation}
For the target fixed-wing ($fw$) we use the x8 Skywalker model from \cite{IROS29}, and simulate the aircraft using JSBSim \cite{IROS30}. For the multi-rotor ($mr$) we provide our own implementation contained within this papers associated repository \footnote{https://github.com/AOS55/Drone2Plane}. Both aircraft have the same 6 degree of freedom 12 state representation form $\mathbf{x}$: 6 translational components $[x, y, z; u, v, w]$ and 6 rotational components $[roll, pitch, yaw; p, q, r]$. Both aircraft have 4 controls $\mathbf{u}$ with normalized inputs for control-deflection and thrust:
\begin{itemize}
        \item The multi-rotor has 4 thrust control commands, one for each prop: $\mathbf{u}^{mr} = [T_{0}, T_{1}, T_{2}, T_{3}]$, where $T_{i}$ is bound from $[0, 1]$.
        \item The fixed-wing airplane is modelled with 3 control surface commands and one throttle command: $\mathbf{u}^{fw}$ = $[ail, elev, tla, rud]$, with bounds: $[[-1, +1], [-1, +1], [0, +1], [-1, +1]]$.
\end{itemize}

\begin{figure}[htb!]
        \centering
        \includegraphics[width=1.0\linewidth]{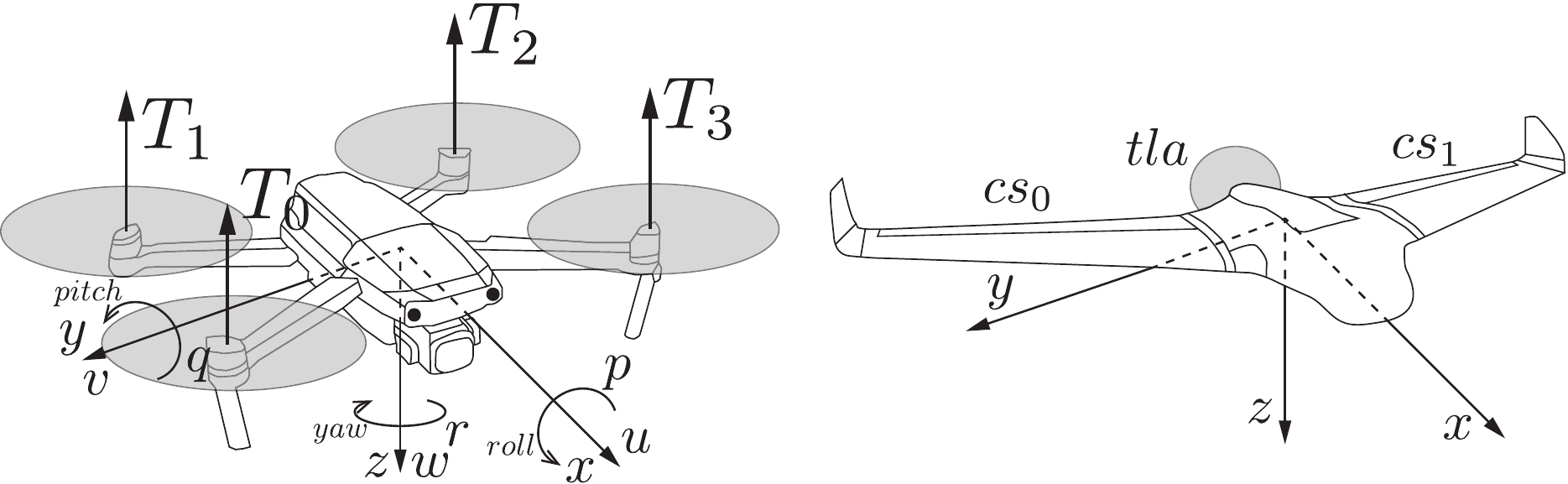}
        \caption{Example platform multi-rotor vehicle and target fixed-wing x8 airplane. Aircraft body axis states are labelled on the multi-rotor.}
        \label{fig:E-Aircraft_Images}
\end{figure}

The actual x8 airplane flown, shown in figure \ref{fig:E-Aircraft_Images}, has 2 full wing control surfaces ($cs$) which are transformed into equivalent elevator \& aileron within the x8 wind tunnel model \cite{IROS29}. The x8's large control surfaces provide lots of control authority in pitch and roll, but the lack of vertical tailplane results in poor directional control and balance. 

Our multi-rotor simulator includes models for drag, motor-thrust and gravitational effects. The vehicle states are updated with a two-step forward euler method using the linear and rotational accelerations generated by these forces. We specified the vehicles performance to satisfy the x8 target platforms required flight envelope, principally by observing when the multi-rotor's motors became saturated when responding to a step-control disturbance on the airplane. This produced a thrust to weight requirement of $\approx 2:1$, performance that is not unrealistic for many small to medium size off-the-shelf quad-copters \cite{IROS42}.

\subsection{MPC}
For our MPC control task we use a state-dependent cost function $\mathbf{q}(\mathbf{x})$ of the form
\begin{equation}
        \mathbf{q}(\mathbf{x}) = \mathbf{w}_{\mathbf{x}}|\mathbf{x}^{fw} - \mathbf{x}^{mr}|^{2}
\end{equation}
Where the state weight matrix $w_{\mathbf{x}}$ has parameters
\begin{equation}
        \mathbf{w}_{\mathbf{x}} = [1, 1, 1, 0, 0, 0, 0, 0, 0, 0, 0, 0] ^{T}
\end{equation}

This corresponds to minimizing the $(x, y, z)$ position error between the fixed-wing and multi-rotor aircraft. This is suitable for higher order fixed-wing control tasks, based on temporal position based objectives \cite{IROS28}. If mimicking rotational vehicle states is desirable, it would not be difficult to include a gimbal controller to directly passthrough the fixed-wing rates from the model. Allowing us to train a vision encoded policy as the attitude of the fixed-wing aircraft has a direct relation to the perspective of the policy. For the control-dependant cost function $\mathbf{p}(\mathbf{u})$ we apply a constant cost weight to all thrust terms of $0.5$. This improves the convergence rate when close to an optimal solution and helps to reduce jittering. To minimize the cost function, $\underset{\mathbf{u}}{\mathrm{argmin}}$ $\mathcal{J}(\mathbf{u}, \mathbf{x}) \ s.t. \ \mathbf{u}_{min} \le \mathbf{u} \le \mathbf{u}_{max}$, we use the Sequential Least Squares Programming (SLSQP) routine from the Python based SciPy optimization library \cite{IROS43} with constraints $[\mathbf{u}_{min}, \mathbf{u}_{max}] \gets [0, 1]$. We use a constant time-horizon $T$ of 1.0s.

\section{EXPERIMENTS}
\label{sec:experiments}
To investigate the limitations of our surrogate learning approach we consider the following procedures:
\begin{itemize}
        \item A disturbance in roll and pitch, to investigate the quad-copter's limitations to mimic the dynamics of the fixed-wing at the limits of the flight envelope. 
        \item A comparison to real-world flight data, to investigate our procedures ability to mimic the aeroplane.
\end{itemize}

\subsection{Disturbance Tracking}

\begin{figure}[htb!]
        \centering
        \includegraphics[width=1.0\linewidth]{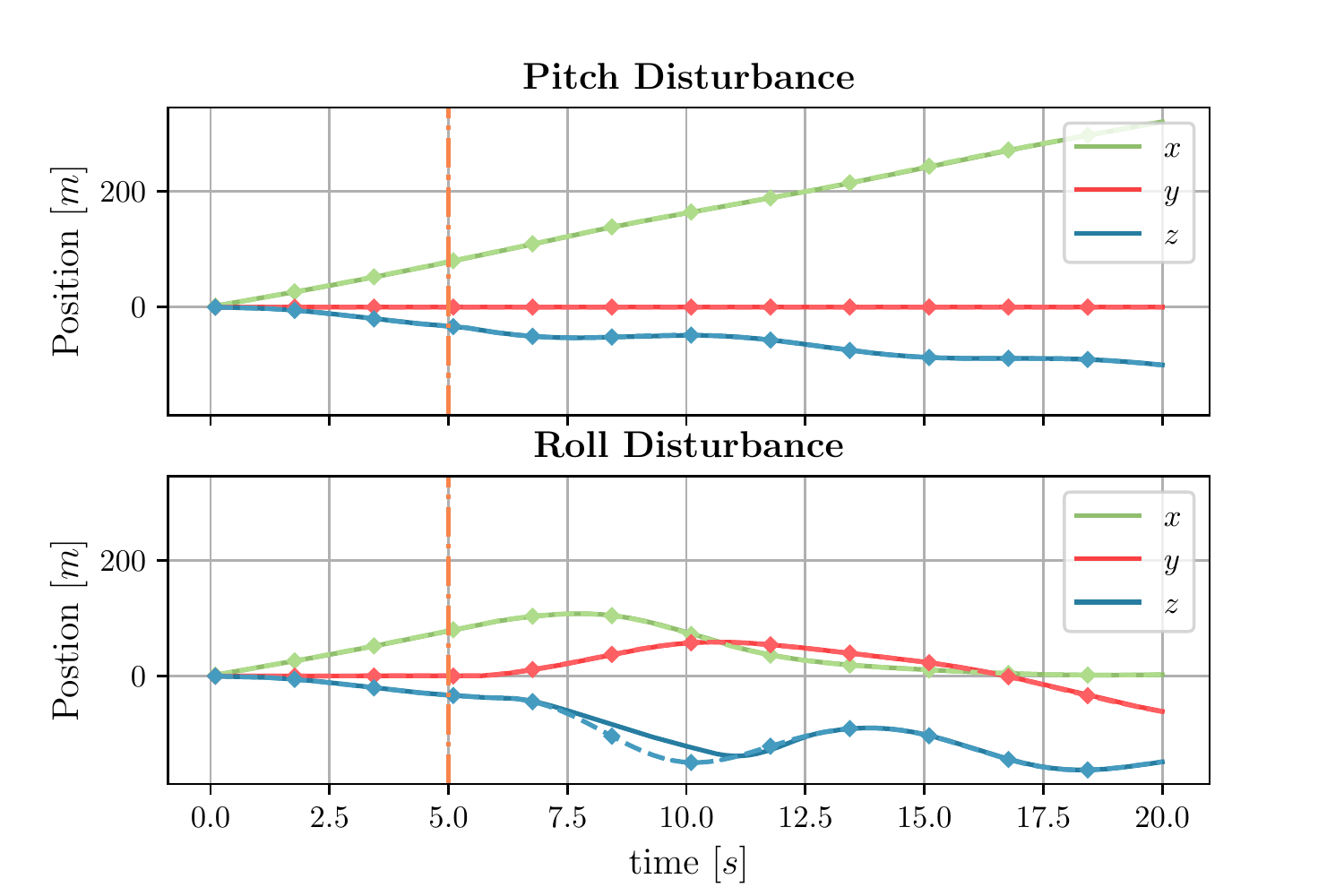}
        \caption{The multi-rotor (solid line) tracks the pitch disturbance well, but becomes saturated when aiming to maintain level with the fixed-wing (dashed-diamond line) following the roll disturbance.}
        \label{fig:B-position_disturbance}
\end{figure}

\begin{figure}[htb!]
        \centering
        \includegraphics[width=1.0\linewidth]{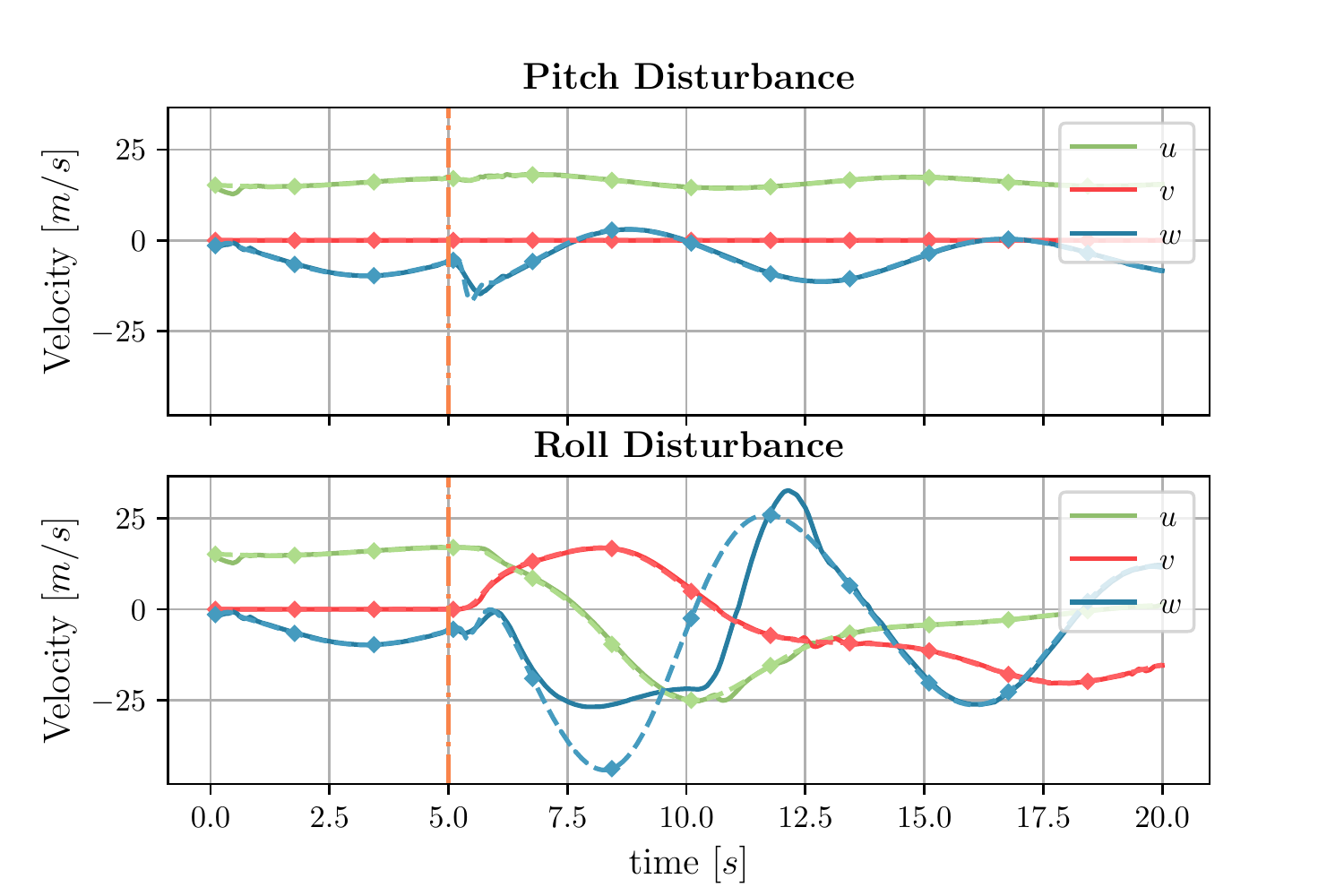}
        \caption{The multi-rotor maintains a similar average speed throughout the manoeuvre in order to minimize the position cost. Note, the multi-rotor does not track linear velocity as part of its cost function.}
        \label{fig:C-velocity_disturbance}
\end{figure}

Figures \ref{fig:B-position_disturbance} \& \ref{fig:C-velocity_disturbance} plot the 3 dimensional position and linear velocity states for the two vehicles respectively. The orange dash-dot vertical line in both figures is the point the disturbance was initiated. For the pitch and roll disturbances we apply an instantaneous up elevator and left aileron deflection for 0.2$s$ respectively. In both cases we use 50 \% of the maximum control deflection for the disturbance. 

Control saturation is shown following the disturbance in roll at 5 seconds on figure \ref{fig:B-position_disturbance} as the quad-copter encounters a slight (10m) departure from tracking in altitude. The system effectively runs out of power required to maintain rate of climb whilst turning. The controller is able to maintain higher power for longer however and recaptures the control by effectively leading the velocity of the fixed-wing aircraft, shown from 9.0s to 12.0s on figure \ref{fig:C-velocity_disturbance}. We find this analysis useful to understand the limitations of our platform vehicle, if tracking closely in this domain was desirable we could increase the performance of our vehicle to maintain position. 

\begin{figure}[htb!]
        \centering
        \includegraphics[width=1.0\linewidth]{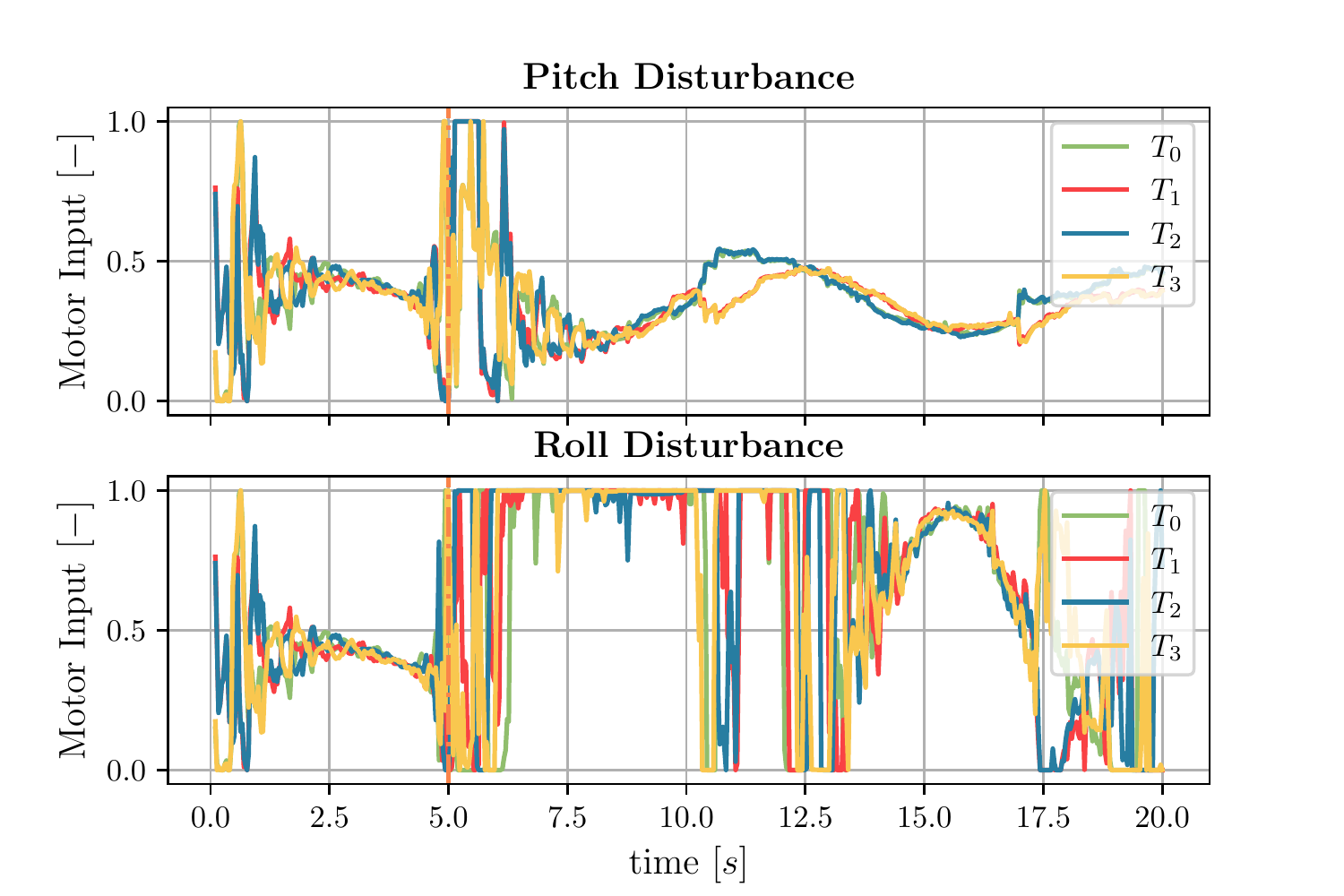}
        \caption{The control input to each of the quad-copter's 4 motors under the pitch and roll disturbances. The roll disturbance is saturated from 5.5s to 10.0s}
        \label{fig:D-control_disturbance}
\end{figure}

Figure \ref{fig:D-control_disturbance} shows the control input required to mimic the disturbances of the fixed wing for each disturbance. The noise at the beginning of each control plot is caused by cold-starting the optimizer, as the control sequence $\tau_{\mathbf{u}}$ starts with all values set to zero. The solution for the optimizer is then progressively improved by warm-starting the optimizer with the last best solution. Once converged to a local optima the variance in control input decays and the noise in the system reduces. Importantly this doesn't cause the aircraft to diverge from the mimicked dynamics, despite following a sub-optimal solution.

\begin{figure}[htb!]
        \centering
        \includegraphics[width=1.0\linewidth]{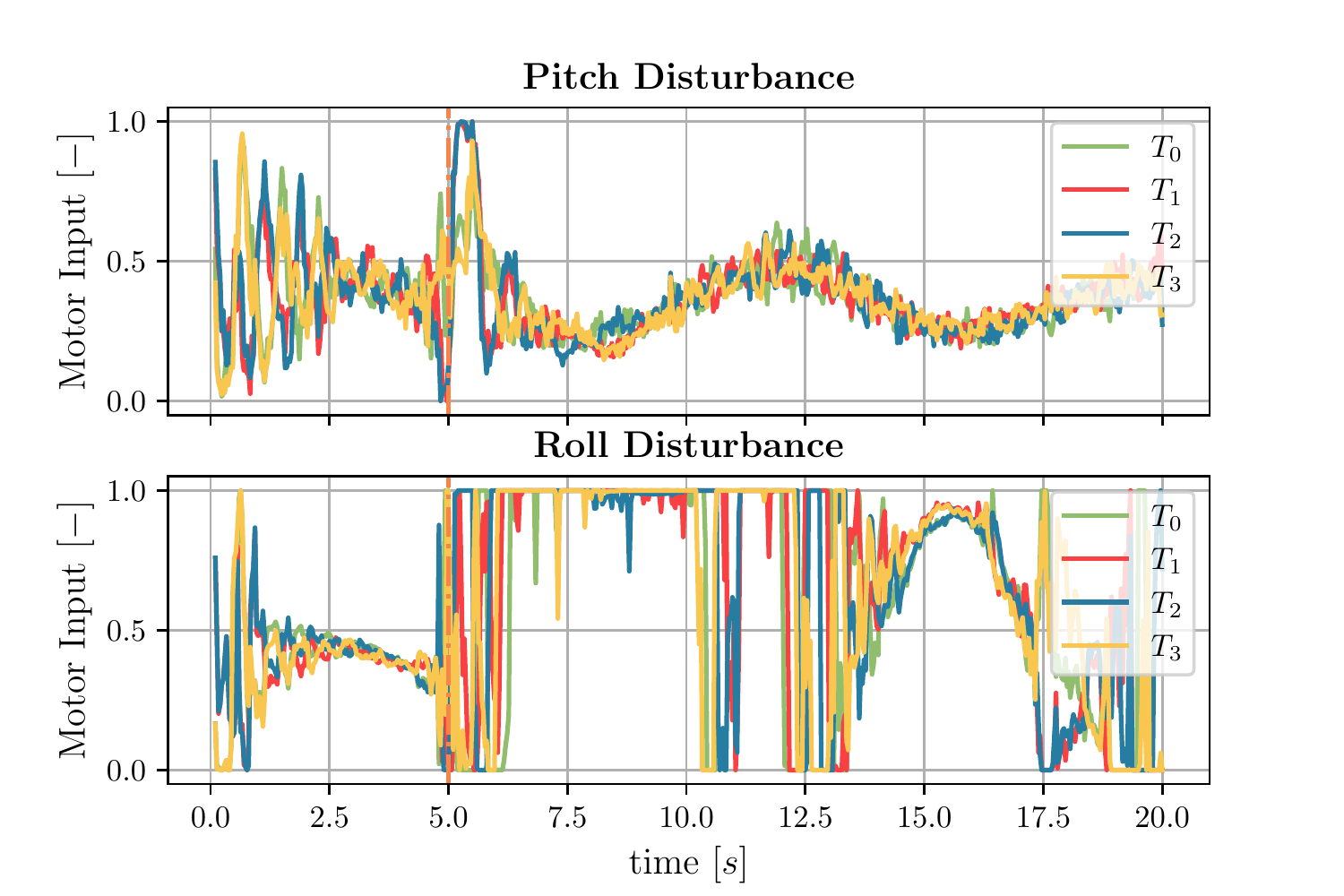}
        \caption{With a $\tau = \frac{1}{30}s$ lag, the controller response is noisier, but tracking accuracy (not shown) is unchanged.}
        \label{fig:E-control_disturbance_lag}
\end{figure}

To simulate the effect of un-modelled dynamics we add a first-order lag filter to the multi-rotor's motors given as

\begin{equation}
        \mathbf{y}_{t} = (1-a) \mathbf{y}_{t-1} + a \mathbf{u}_{t} \ \ \ a = \frac{\delta_{t}}{\delta_{t} + T_{lag}}
\end{equation}

Where $\mathbf{y_{t}}$ is the command sent to the linear motor model at time $t$, $\delta_{t}$ is the step-size of the simulation, and $T_{lag} = \frac{1}{30}s$ is the size of the lag. But, we do not include the lag model within the MPC model. We found this did not cause any significant effects on the ability of the quad-copter to track the commanded signal but, as expected, slightly increases the noise of the motor input command, shown in figure \ref{fig:E-control_disturbance_lag}.  

\subsection{Real-World Comparison}
\begin{figure}[htb!]
        \centering
        \includegraphics[width=1.0\linewidth]{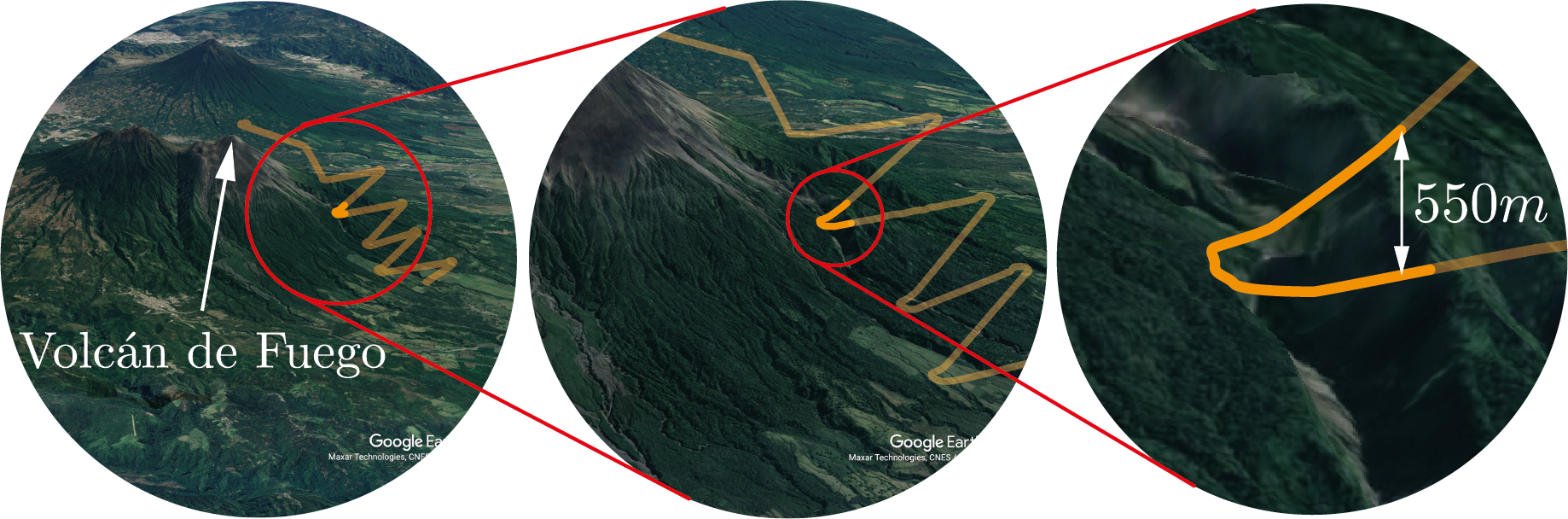}
        \caption{Climbing Turn section starting at 15,000', around Volc\'an de Fuego. Solid orange segment indicates flight track used over the 60 second period.}
        \label{fig:G-3d_guatemala}
\end{figure}

To validate our approach on a real-world dataset we use our algorithm to mimic the motion of an x8 on a gas-sensing mission to the summit of Volc\'an de Fuego in Guatemala using the same platform drone. We consider a demanding 60 second climbing turn section, shown in figure \ref{fig:G-3d_guatemala}.

\begin{figure}[htb!]
        \centering
        \includegraphics[width=1.0\linewidth]{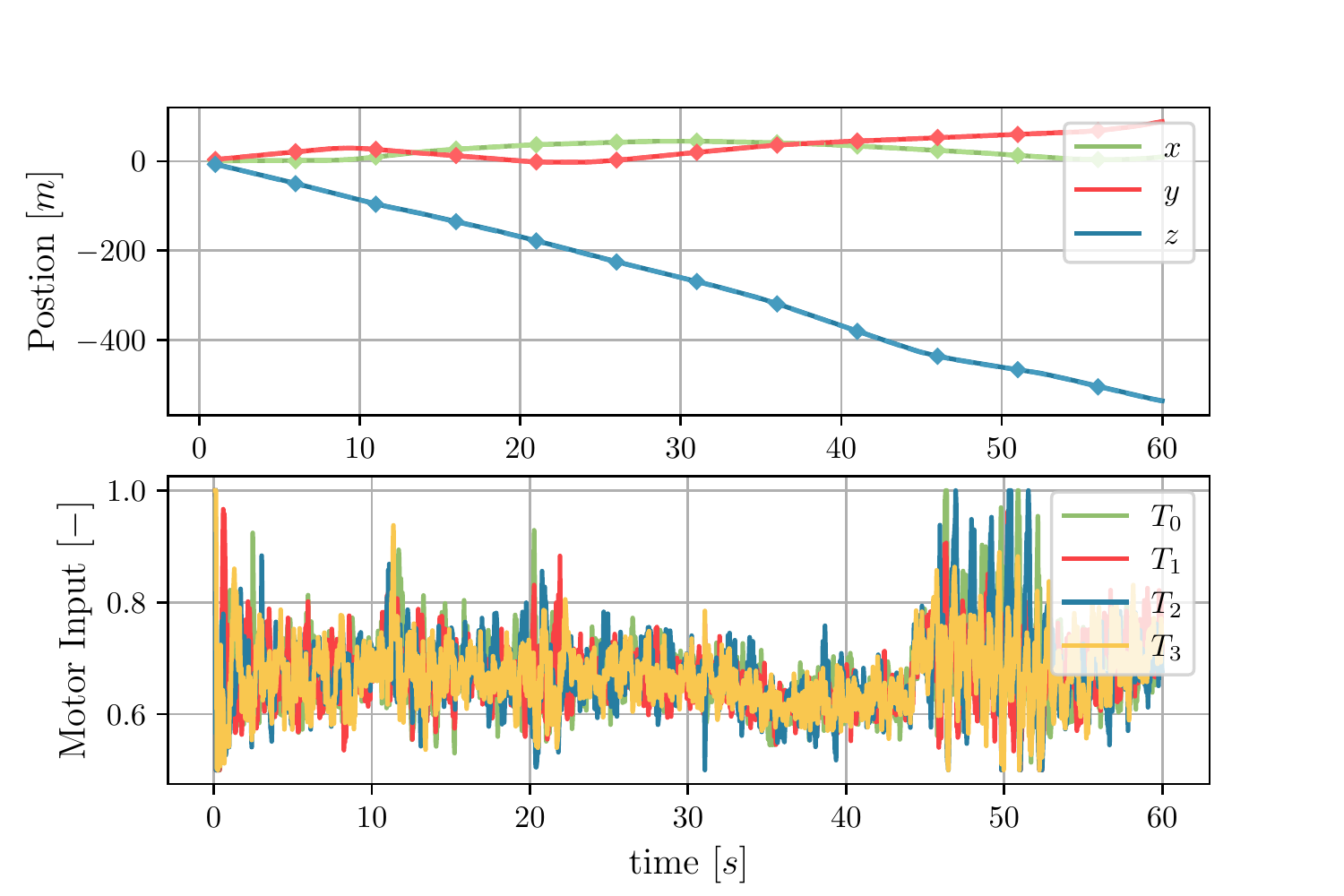}
        \caption{Position during manoeuvre in figure \ref{fig:G-3d_guatemala}, aircraft dashed, multi-rotor solid, showing little to no error. Lower plot shows the control commands required to track the manoeuvre.}
        \label{fig:H-thr_pos}
\end{figure}

We find the multi-rotor platform is capable of tracking the fixed-wing demonstration with little to no error through the whole manoeuvre with relatively small control noise, shown in figure \ref{fig:H-thr_pos}. Interestingly the controller approaches saturation during the level out from the turn, where it needs to turn aggressively and continue the climb. 

We find that the trajectory tracking error of the real world data in figure \ref{fig:H-thr_pos} is relatively small, with mean squared error in $[x, y, z]$ positions of $[0.00182, 0.00357, 0.310]$ respectively. 


\section{FUTURE WORK \& LIMITATIONS}

A key limitation of our approach is the time taken to calculate trajectories for our MPC loop. Even in the best case the frequency of our controller to update on modern hardware is 10 to 100 times slower than is required to run online. A large constraint to this is the need to use a sequential quadratic programming solver, as the core algorithm is fundamentally serial and the computational expense to calculate gradients and simulate model state roll-outs is significant. Information Theoretic MPC methods such as MPPI \cite{IROS39} offer a solution to this challenge, via a sampling based MPC procedure, based on KL divergence and free energy. This allows us to take full advantage of parallel compute and does not require gradients to be calculated for the optimization procedure. Initially, we found that running MPPI, with the quad-copter simulator on parallel CPU compute, provided insufficient samples to converge to a stable solution. We therefore aimed to use a neural-network to represent our model dynamics, allowing us to take advantage of accelerated parallel GPU compute and collect many more samples faster. To train our network we investigated bootstrapping from demonstrations collected with both random actions and an MPC oracle along with simply training within the MPPI framework. A unique challenge that stems from the quad-copters fundamental instability is that random actions frequently lead to the aircraft rotating into an uncontrolled state. The original MPPI paper \cite{IROS39} solves this on the quad-copter task by including angular rates as part of the control input. We found that whilst position and linear/angular velocities can be predicted with small error, angles could only be learned to within $1^{c}$ of error, which is far too large of an error to use with MPPI.

More recent Model-Based RL algorithms such as MBPO \cite{IROS44} or PETS \cite{IROS45} may offer a better solution to solving our control problem, with their CMA-ES ensemble maximum likelihood \cite{IROS46} based model learning procedure. Alternatively, we may simply need to collect more demonstrations for supervised training, either from the MPC oracle or human demonstrations. 

Once this computational bottleneck is overcome we aim to experiment with the controller in the real-world. Such as: online episodic reinforcement learning and safety constrained RL tasks, for example flying at high speed around a car-park or navigating to landing spots in built-up areas.









\bibliography{Drone2Plane} 

\begin{thebibliography}{10}

\bibitem{IROS1}
V.~Mnih, K.~Kavukcuoglu, D.~Silver, A.~Graves, I.~Antonoglou, D.~Wierstra, and
  M.~Riedmiller, ``Playing atari with deep reinforcement learning,'' 2013.

\bibitem{IROS2}
J.~Schulman, S.~Levine, P.~Moritz, M.~I. Jordan, and P.~Abbeel, ``Trust
  {{Region Policy Optimization}},'' {\em arXiv:1502.05477 [cs]}, Apr. 2017.

\bibitem{IROS3}
OpenAI, M.~Andrychowicz, B.~Baker, M.~Chociej, R.~Jozefowicz, B.~McGrew,
  J.~Pachocki, A.~Petron, M.~Plappert, G.~Powell, A.~Ray, J.~Schneider,
  S.~Sidor, J.~Tobin, P.~Welinder, L.~Weng, and W.~Zaremba, ``Learning
  dexterous in-hand manipulation,'' 2019.

\bibitem{IROS4}
S.~Levine, C.~Finn, T.~Darrell, and P.~Abbeel, ``End-to-end training of deep
  visuomotor policies,'' {\em J. Mach. Learn. Res.}, vol.~17, p.~1334–1373,
  jan 2016.

\bibitem{IROS41}
H.~Zhu, J.~Yu, A.~Gupta, D.~Shah, K.~Hartikainen, A.~Singh, V.~Kumar, and
  S.~Levine, ``The ingredients of real world robotic reinforcement learning,''
  in {\em International Conference on Learning Representations}, 2020.

\bibitem{IROS5}
A.~X. Lee, C.~M. Devin, Y.~Zhou, T.~Lampe, K.~Bousmalis, J.~T. Springenberg,
  A.~Byravan, A.~Abdolmaleki, N.~Gileadi, D.~Khosid, C.~Fantacci, J.~E. Chen,
  A.~Raju, R.~Jeong, M.~Neunert, A.~Laurens, S.~Saliceti, F.~Casarini,
  M.~Riedmiller, raia hadsell, and F.~Nori, ``Beyond pick-and-place: Tackling
  robotic stacking of diverse shapes,'' in {\em 5th Annual Conference on Robot
  Learning}, 2021.

\bibitem{IROS6}
R.~J. Clarke, L.~Fletcher, C.~Greatwood, A.~Waldock, and T.~S. Richardson,
  ``Closed-{{Loop Q-Learning Control}} of a {{Small Unmanned Aircraft}},'' in
  {\em {{AIAA Scitech}} 2020 {{Forum}}}, ({Orlando, FL}), {American Institute
  of Aeronautics and Astronautics}, Jan. 2020.

\bibitem{IROS47}
L.~Fletcher, R.~Clarke, T.~Richardson, and M.~Hansen, ``Reinforcement learning
  for a perched landing in the presence of wind,'' in {\em AIAA Scitech 2021
  Forum}, American Institution of Aeronautics and Astronautics, Jan. 2021.
\newblock 2021 AIAA SciTech Forum ; Conference date: 04-01-2021 Through
  15-01-2021.

\bibitem{IROS7}
E.~Bohn, E.~M. Coates, S.~Moe, and T.~A. Johansen, ``Deep reinforcement
  learning attitude control of fixed-wing uavs using proximal policy
  optimization,'' {\em 2019 International Conference on Unmanned Aircraft
  Systems (ICUAS)}, Jun 2019.

\bibitem{IROS34}
X.~Ding, P.~Guo, K.~Xu, and Y.~Yu, ``A review of aerial manipulation of
  small-scale rotorcraft unmanned robotic systems,'' {\em Chinese Journal of
  Aeronautics}, vol.~32, pp.~200--214, Jan. 2019.

\bibitem{IROS36}
A.~Loquercio, A.~I. Maqueda, C.~R. del Blanco, and D.~Scaramuzza, ``Dronet:
  Learning to fly by driving,'' {\em IEEE Robotics and Automation Letters},
  vol.~3, no.~2, pp.~1088--1095, 2018.

\bibitem{IROS35}
E.~Kaufmann, A.~Loquercio, R.~Ranftl, A.~Dosovitskiy, V.~Koltun, and
  D.~Scaramuzza, ``Deep drone racing: Learning agile flight in dynamic
  environments,'' in {\em CoRL}, 2018.

\bibitem{IROS8}
B.~Eysenbach, S.~Gu, J.~Ibarz, and S.~Levine, ``Leave no trace: Learning to
  reset for safe and autonomous reinforcement learning,'' in {\em International
  Conference on Learning Representations}, 2018.

\bibitem{IROS9}
J.~Ibarz, J.~Tan, C.~Finn, M.~Kalakrishnan, P.~Pastor, and S.~Levine, ``How to
  train your robot with deep reinforcement learning: Lessons we have learned,''
  {\em The International Journal of Robotics Research}, vol.~40, pp.~698--721,
  Apr. 2021.

\bibitem{IROS10}
J.~V. Foster and D.~Hartman, ``High-{{Fidelity Multi-Rotor Unmanned Aircraft
  System}} ({{UAS}}) {{Simulation Development}} for {{Trajectory Prediction
  Under Off-Nominal Flight Dynamics}},'' in {\em 17th {{AIAA Aviation
  Technology}}, {{Integration}}, and {{Operations Conference}}}, ({Denver,
  Colorado}), {American Institute of Aeronautics and Astronautics}, June 2017.

\bibitem{IROS19}
K.~P. Srinivasan, B.~Eysenbach, S.~Ha, J.~Tan, and C.~Finn, ``Learning to be
  safe: Deep rl with a safety critic,'' {\em ArXiv}, vol.~abs/2010.14603, 2020.

\bibitem{IROS18}
S.~Thrun, {\em Lifelong Learning Algorithms}, p.~181–209.
\newblock USA: Kluwer Academic Publishers, 1998.

\bibitem{IROS37}
DeepMind, ``https://deepmind.com/blog/announcements/mujoco.''
\newblock 2022-02-14.

\bibitem{IROS20}
M.~E. Taylor and P.~Stone, ``Transfer learning for reinforcement learning
  domains: A survey,'' {\em J. Mach. Learn. Res.}, vol.~10, p.~1633–1685, dec
  2009.

\bibitem{IROS21}
S.~J. Pan and Q.~Yang, ``A survey on transfer learning,'' {\em IEEE
  Transactions on Knowledge and Data Engineering}, vol.~22, no.~10,
  pp.~1345--1359, 2010.

\bibitem{IROS22}
C.~Tan, F.~Sun, T.~Kong, W.~Zhang, C.~Yang, and C.~Liu, ``A survey on deep
  transfer learning,'' 2018.
\newblock cite arxiv:1808.01974Comment: The 27th International Conference on
  Artificial Neural Networks (ICANN 2018).

\bibitem{IROS11}
J.~Tobin, R.~Fong, A.~Ray, J.~Schneider, W.~Zaremba, and P.~Abbeel, ``Domain
  randomization for transferring deep neural networks from simulation to the
  real world,'' {\em 2017 IEEE/RSJ International Conference on Intelligent
  Robots and Systems (IROS)}, pp.~23--30, 2017.

\bibitem{IROS15}
F.~Sadeghi and S.~Levine, ``Cad2rl: Real single-image flight without a single
  real image,'' in {\em 13th converences on Robotics: Science and Systems},
  pp.~34--44, 07 2017.

\bibitem{IROS14}
F.~Muratore, C.~Eilers, M.~Gienger, and J.~Peters, ``Data-efficient domain
  randomization with bayesian optimization,'' {\em IEEE Robotics and Automation
  Letters}, vol.~6, pp.~911--918, 2021.

\bibitem{IROS40}
E.~Valassakis, Z.~Ding, and E.~Johns, ``Crossing the gap: A deep dive into
  zero-shot sim-to-real transfer for dynamics,'' {\em 2020 IEEE/RSJ
  International Conference on Intelligent Robots and Systems (IROS)},
  pp.~5372--5379, 2020.

\bibitem{IROS17}
M.~Hazara and V.~Kyrki, ``Transferring generalizable motor primitives from
  simulation to real world,'' {\em IEEE Robotics and Automation Letters},
  vol.~4, no.~2, pp.~2172--2179, 2019.

\bibitem{IROS12}
D.~Gandhi, L.~Pinto, and A.~K. Gupta, ``Learning to fly by crashing,'' {\em
  2017 IEEE/RSJ International Conference on Intelligent Robots and Systems
  (IROS)}, pp.~3948--3955, 2017.

\bibitem{IROS23}
Y.~Lu, K.~Hausman, Y.~Chebotar, M.~Yan, E.~Jang, A.~Herzog, T.~Xiao, A.~Irpan,
  M.~Khansari, D.~Kalashnikov, and S.~Levine, ``{AW}-opt: Learning robotic
  skills with imitation andreinforcement at scale,'' in {\em 5th Annual
  Conference on Robot Learning}, 2021.

\bibitem{IROS24}
R.~C. Julian, B.~Swanson, G.~S. Sukhatme, S.~Levine, C.~Finn, and K.~Hausman,
  ``Never stop learning: The effectiveness of fine-tuning in robotic
  reinforcement learning,'' in {\em CoRL}, 2020.

\bibitem{IROS16}
C.~Sun, J.~Orbik, C.~M. Devin, B.~H. Yang, A.~Gupta, G.~Berseth, and S.~Levine,
  ``Fully autonomous real-world reinforcement learning with applications to
  mobile manipulation,'' in {\em 5th Annual Conference on Robot Learning},
  2021.

\bibitem{IROS32}
Calspan, ``https://calspan.com/aerospace/advanced-flight-test-training.''
\newblock 2022-02-14.

\bibitem{IROS31}
R.~P. Harper and G.~E. Cooper, ``Handling qualities and pilot evaluation,''
  {\em Journal of Guidance, Control, and Dynamics}, vol.~9, pp.~515--529, Sept.
  1986.

\bibitem{IROS25}
M.~Shafer, ``In-flight simulation studies at the nasa dryden flight research
  facility,'' {\em Scientific and Technical Information Program}, 1992.

\bibitem{IROS26}
R.~S. Edmonson and J.~Kemper, ``Adaptive flight control systems on calspan
  learjet,'' in {\em AIAA Scitech 2019 Forum}, (San Diego, California),
  American Institute of Aeronautics and Astronautics, Jan. 2019.

\bibitem{IROS33}
M.~E. Knapp, T.~Berger, M.~Tischler, and M.~C. Cotting, ``Development of a
  {{Full Envelope Flight Identified F-16 Simulation Model}},'' in {\em 2018
  {{AIAA Atmospheric Flight Mechanics Conference}}}, ({Kissimmee, Florida}),
  {American Institute of Aeronautics and Astronautics}, Jan. 2018.

\bibitem{IROS38}
D.~Q. Mayne, ``Model predictive control: Recent developments and future
  promise,'' {\em Automatica}, vol.~50, no.~12, pp.~2967--2986, 2014.

\bibitem{IROS49}
K.~P. Wabersich and M.~N. Zeilinger, ``Safe exploration of nonlinear dynamical
  systems: A predictive safety filter for reinforcement learning,'' {\em
  ArXiv}, vol.~abs/1812.05506, 2018.

\bibitem{IROS27}
T.~Erez, K.~Lowrey, Y.~Tassa, V.~Kumar, S.~Kolev, and E.~Todorov, ``An
  integrated system for real-time model predictive control of humanoid
  robots,'' in {\em 2013 13th IEEE-RAS International Conference on Humanoid
  Robots (Humanoids)}, pp.~292--299, 2013.

\bibitem{IROS48}
S.~Gros, M.~Zanon, R.~Quirynen, A.~Bemporad, and M.~Diehl, ``From linear to
  nonlinear mpc: bridging the gap via the real-time iteration,'' {\em
  International Journal of Control}, vol.~93, pp.~62 -- 80, 2020.

\bibitem{IROS29}
K.~Gryte, R.~Hann, M.~Alam, J.~Rohac, T.~Johansen, and T.~Fossen, ``Aerodynamic
  modeling of the skywalker x8 fixed-wing unmanned aerial vehicle,''
  pp.~826--835, 06 2018.

\bibitem{IROS30}
M.~Madden, {\em Architecting a Simulation Framework for Model Rehosting}.

\bibitem{IROS42}
DJI, ``https://www.dji.com/uk/dji-fpv/specs.''
\newblock 2022-02-24.

\bibitem{IROS28}
ICAO, {\em Procedures for Air Navigation Services, Air Traffic Management},
  pp.~142--164.
\newblock 999 Robert-Bourassa Boulevard, Montréal, Quebec, Canada H3C 5H7:
  International Civil Aviation Organization, 2016.

\bibitem{IROS43}
P.~Virtanen, R.~Gommers, T.~E. Oliphant, M.~Haberland, T.~Reddy, D.~Cournapeau,
  E.~Burovski, P.~Peterson, W.~Weckesser, J.~Bright, S.~J. {van der Walt},
  M.~Brett, J.~Wilson, K.~J. Millman, N.~Mayorov, A.~R.~J. Nelson, E.~Jones,
  R.~Kern, E.~Larson, C.~J. Carey, {\.I}.~Polat, Y.~Feng, E.~W. Moore,
  J.~{VanderPlas}, D.~Laxalde, J.~Perktold, R.~Cimrman, I.~Henriksen, E.~A.
  Quintero, C.~R. Harris, A.~M. Archibald, A.~H. Ribeiro, F.~Pedregosa, P.~{van
  Mulbregt}, and {SciPy 1.0 Contributors}, ``{{SciPy} 1.0: Fundamental
  Algorithms for Scientific Computing in Python},'' {\em Nature Methods},
  vol.~17, pp.~261--272, 2020.

\bibitem{IROS39}
G.~Williams, N.~Wagener, B.~Goldfain, P.~Drews, J.~M. Rehg, B.~Boots, and E.~A.
  Theodorou, ``Information theoretic mpc for model-based reinforcement
  learning,'' in {\em 2017 IEEE International Conference on Robotics and
  Automation (ICRA)}, pp.~1714--1721, 2017.

\bibitem{IROS44}
M.~Janner, J.~Fu, M.~Zhang, and S.~Levine, ``When to trust your model:
  Model-based policy optimization,'' in {\em Advances in Neural Information
  Processing Systems} (H.~Wallach, H.~Larochelle, A.~Beygelzimer,
  F.~d\textquotesingle Alch\'{e}-Buc, E.~Fox, and R.~Garnett, eds.), vol.~32,
  Curran Associates, Inc., 2019.

\bibitem{IROS45}
K.~Chua, R.~Calandra, R.~McAllister, and S.~Levine, ``Deep reinforcement
  learning in a handful of trials using probabilistic dynamics models,'' in
  {\em Advances in Neural Information Processing Systems} (S.~Bengio,
  H.~Wallach, H.~Larochelle, K.~Grauman, N.~Cesa-Bianchi, and R.~Garnett,
  eds.), vol.~31, Curran Associates, Inc., 2018.

\bibitem{IROS46}
N.~Hansen and A.~Ostermeier, ``Completely derandomized self-adaptation in
  evolution strategies,'' {\em Evolutionary Computation}, vol.~9, no.~2,
  pp.~159--195, 2001.

\end{thebibliography}
\bibliographystyle{ieeetr}
        

\end{document}